\documentclass[conference]{IEEEtran}
\IEEEoverridecommandlockouts
\usepackage{multirow}
\usepackage{booktabs} 
\usepackage{url} 
\usepackage{hyperref} 
\usepackage{cite}
\usepackage{amsmath,amssymb,amsfonts}
\usepackage{algorithmic}
\usepackage{graphicx}
\usepackage{textcomp}
\usepackage{xcolor}
\usepackage{ragged2e}

\def\BibTeX{{\rm B\kern-.05em{\sc i\kern-.025em b}\kern-.08em
    T\kern-.1667em\lower.7ex\hbox{E}\kern-.125emX}}

\author{
  \IEEEauthorblockN{%
    \centering
    Jun Yu\textsuperscript{1,2}, WenJian Wang\textsuperscript{3,*}
  }
  \IEEEauthorblockA{%
    \parbox[t]{\textwidth}{%
      \raggedright
      \textit{1 School of Computer and Information Technology (School of Big Data), Shanxi University, Taiyuan, China}
    }
  }
  \IEEEauthorblockA{%
    \parbox[t]{\textwidth}{%
      \raggedright
      \textit{2 Taihang Laboratory In Shanxi Province (Advanced Computing Laboratory In Shanxi Province), Taiyuan, China}
    }
  }
  \IEEEauthorblockA{%
    \parbox[t]{\textwidth}{%
      \raggedright
      \textit{3 Key Laboratory of Computational Intelligence and Chinese Information Processing of Ministry of Education, Shanxi University, Taiyuan, China}
    }
  }
  \IEEEauthorblockA{%
    \centering
    \textsuperscript{*} Correspondence: y.j@sxu.edu.cn, wjwang@sxu.edu.cn
  }
}

\title{DDNet: Deformable Convolution and Dense FPN for Surface Defect Detection in Recycled Books}

\begin{document}

\begin{minipage}{\textwidth}

\maketitle
\end{minipage}

\begin{abstract}
Recycled and recirculated books, such as ancient texts and reused textbooks, hold significant value in the second-hand goods market, with their worth largely dependent on surface preservation. However, accurately assessing surface defects is challenging due to the wide variations in shape, size, and the often imprecise detection of defects. To address these issues, we propose DDNet, an innovative detection model designed to enhance defect localization and classification.
DDNet introduces a surface defect feature extraction module based on a deformable convolution operator (DC) and a densely connected FPN module (DFPN). The DC module dynamically adjusts the convolution grid to better align with object contours, capturing subtle shape variations and improving boundary delineation and prediction accuracy. Meanwhile, DFPN leverages dense skip connections to enhance feature fusion, constructing a hierarchical structure that generates multi-resolution, high-fidelity feature maps, thus effectively detecting defects of various sizes.
In addition to the model, we present a comprehensive dataset specifically curated for surface defect detection in recycled and recirculated books. This dataset encompasses a diverse range of defect types, shapes, and sizes, making it ideal for evaluating the robustness and effectiveness of defect detection models.
Through extensive evaluations, DDNet achieves precise localization and classification of surface defects, recording a mAP value of 46.7\% on our proprietary dataset—an improvement of 14.2\% over the baseline model—demonstrating its superior detection capabilities.
\end{abstract}


\section{Introduction}
The market for recycled and recirculated books \cite{b1, b2} has gained significant popularity due to its diverse collections and unique value. These books not only carry economic significance but also play a crucial role in the dissemination of knowledge and the preservation of cultural heritage. However, beyond their content, the value of these books is heavily influenced by their physical condition. Traditional manual inspection methods, being both inefficient and subjective, struggle to meet the growing market demand. To address these challenges, this study proposes the use of computer vision technology not only to automatically detect surface defects in second-hand books but also to incorporate re-identification techniques \cite{b3, b4, b5}. By implementing re-identification, the system can accurately match and track individual books across various stages of circulation, ensuring consistent quality assessments and enhancing both detection accuracy and operational efficiency.

\begin{figure}[t]
    \vspace{-0.5cm}
\centerline{
\includegraphics[trim={0 7cm 0 0}, 
        clip,
        width=1\linewidth]{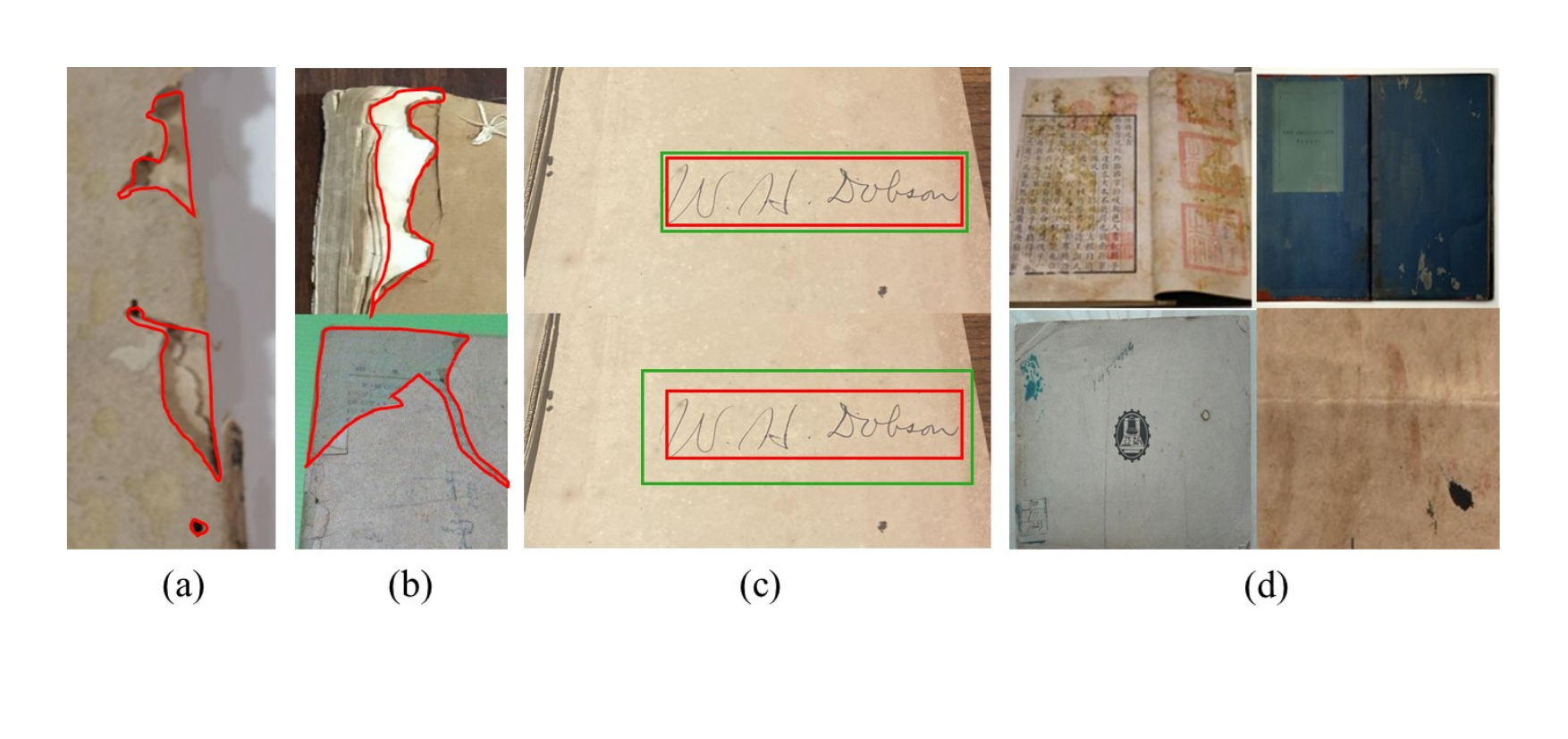}}
 \vspace{-1cm} 
\caption{Typical defects on the surface of recovering and re-circulating books.  (a) Varying defect sizes; (b) Varying defect shapes; (c) Defect location accuracy. The red lines indicate the ground truth positions, while the green lines represent the predicted locations; (d) Typical defect images.
} \vspace{-0.7cm} 
\label{fig:F1}
\end{figure}

\begin{figure*}[t]
\centerline{
\includegraphics[trim={0 13cm 0 0}, 
        clip,width=1\linewidth]{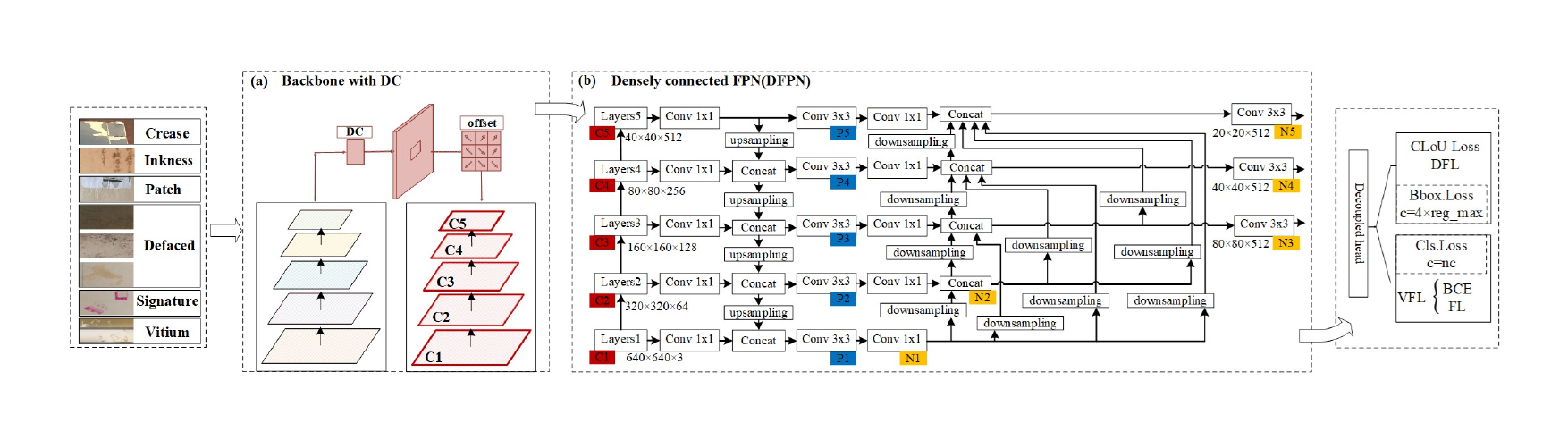}}
         \vspace{-0.7cm} 
\caption{Proposed DDNet model framework: (a) Feature extraction with DC; (b) Dense feature fusion pyramid. }
         \vspace{-0.5cm} 
\label{fig:F2}
\end{figure*}

Defect detection is essential in modern industrial production and quality control. Machine vision, with its non-contact, non-destructive nature and ability to operate efficiently under harsh conditions, is increasingly important in automated production and intelligent pricing. Surface defect detection can be categorized into traditional methods and deep learning-based approaches. Traditional methods, including transform domain analysis \cite{b6}, statistical techniques \cite{b7,b8}, and machine learning \cite{b9, b10, b11}, rely on image processing and manually designed features. While effective in simple environments, these methods often struggle with complex scenarios. In contrast, deep learning has significantly improved defect detection. Deep learning models automatically extract features, enhancing accuracy and generalization for diverse defects. These methods are typically divided into two categories: two-stage networks, like Faster R-CNN, which offer higher accuracy but are computationally intensive, and single-stage networks, like YOLO and SSD, which are more efficient and suitable for real-time applications.

Recently, the application of diffusion models \cite{b17, b18} has further advanced defect detection capabilities. Diffusion models, with their iterative approach to refining images through noise addition and removal, can generate highly detailed and accurate representations of defects. This method not only improves the model's ability to detect subtle and complex defects but also enhances the robustness of the detection process under varying conditions. By combining diffusion models with traditional deep learning techniques, the defect detection process can achieve unprecedented levels of accuracy and efficiency, making it a powerful tool in modern quality control.

Despite the progress, the automatic detection of surface defects in recycled books remains underexplored. As shown in Fig.~\ref{fig:F1}, the unique challenges faced by this study include significant variations in defect shapes, which standard convolutional layers and fixed kernels may not be able to effectively adapt to, large size differences between and within defect types, and precise location of defects.

To address the challenges of surface defect detection in recycled and recirculated books, we propose a high-quality defect detection framework DDNet, which extends the YOLOv8 architecture with two carefully designed modules: a feature extraction module fused with a deformable convolution operator (DC) and a densely connected feature pyramid module (DFPN).

The main contributions of this paper include:

\begin{itemize}
    \item We introduce a deep learning-based approach designed to address the challenge of surface defect detection in recycled and recirculated books. This method enables fast, automatic, and real-time identification of defects in second-hand materials. Comparisons with widely-used detection algorithms, such as YOLOv8, the YOLO series, and Rtm-Net, demonstrate that our model achieves superior performance in terms of mAP.

    \item To effectively manage the irregularities in image shape and scale, we incorporate the DC and DFPN modules. These modules are specifically designed to handle variable image distortions, providing robust solutions for practical scenarios in surface defect detection.

    \item A comprehensive dataset of 6,366 images is constructed, featuring surface defects from a variety of recycled and recirculated books, including ancient manuscripts and modern textbooks. The dataset includes annotations for six types of surface defects, such as stains, insect damage, and general wear and tear.
\end{itemize}

\section{Proposed Method}

\begin{table*}[t]
    \centering
    \vspace{-0.25cm} 
\caption{\textbf{Inspection results (\%) on DDNet. }}
    \vspace{-0.3cm} 
        \begin{tabular}{|c|c|c|c|c|c|c|c|} \hline 
         \textbf{Models} & \textbf{mAP} & \textbf{Inkiness} & \textbf{Vitium} & \textbf{Crease} & \textbf{Defaced} & \textbf{Patch} & \textbf{Signature}\\\hline 
 YOLOv5\cite{b12}& 7.7& 1.4& 1.3& 2.8& 1.1& 1.1&31.2\\
 YOLOv6\cite{b13}& 26.2& 16.8& 17.1& 9.8& 18.2& 19.2&76.0\\
 YOLOX\cite{b15}& 27.7& 21.4& 17.0& 13.4	& 21.7& 21.4&70.8\\
 rtm\cite{b16}& 28.4& 19.0& 15.6& 10.5& 22.0& 37.5&65.6\\ 
          \textbf{YOLOv8 (baseline)\cite{b14} }&   32.5&  20.9&  15.1&  12.4&  17.4&  19.4&  75.6\\ \hline 
         YOLOV8 +DC&   46.0&  43.5&  22.7&  \textbf{28.3}&  34.9&  51.7&   \textbf{94.8}\\ 
         YOLOV8 + DFPN&   45.5&  42.9&  24.6&  27.5&  31.5&  52.7&  94.0\\\hline 
         \textbf{DDNet (Ours)}&  \textbf{46.7}&  \textbf{46.1}&  \textbf{25.8}&  23.6&  \textbf{37.7}&  \textbf{55.2}&  91.6\\\hline
    \end{tabular}
    \label{tab:1}
        \vspace{-0.5cm} 
\end{table*}

\begin{table*}[t]
    \centering
         \caption{\textbf{Results (\%) of Different Models with DC-Added CSPDarkNet.}}
    \vspace{-0.3cm} 
    \begin{tabular}{|c|c|c|c|c|c|c|c|c|c|c|} \hline 
\multicolumn{4}{|c|}{\textbf{CSPDarkNet stages with DC added}} & \multirow{2}{*}{\textbf{mAP}} & \multirow{2}{*}{\textbf{Inkiness}} & \multirow{2}{*}{\textbf{Vitium}} & \multirow{2}{*}{\textbf{Crease}} & \multirow{2}{*}{\textbf{Defaced}} & \multirow{2}{*}{\textbf{Patch}} & \multirow{2}{*}{\textbf{Signature}} \\
\cline{1-4}
\textbf{stage2} & \textbf{stage3} & \textbf{stage4} & \textbf{stage5} & & & & & & & \\
\hline  
        $\checkmark$ & $\checkmark$ & $\checkmark$ & $\checkmark$ 
&  42.2&  41.4&  24.3&  25.8&  32.7&  36.9&
92.0\\ 
         $\times$ & $\checkmark$ & $\checkmark$ & $\checkmark$ 
&  45.0&  43.6&  25.1&  23.6&  34.9&  47.5&95.5\\
        $\times$ & $\times$ & $\checkmark$ & $\checkmark$ 
&  \textbf{47.3} &  \textbf{43.5} &\textbf{30.6} &\textbf{28.3}& \textbf{34.9} &  \textbf{51.7} & \textbf{94.8}\\  
        $\times$ & $\times$ & $\times$ & $\checkmark$ 
&  45.8&  46.0&  25.5&  25.9&  28.5&  53.4&95.7\\ \hline
    \end{tabular}
 
    \label{tab:2}
            \vspace{-0.5cm} 
\end{table*}

\subsection{System Overview  }
The DDNet framework, as illustrated in Fig.~\ref{fig:F2}, comprises four key components: the feature extraction network, the DC module, the DFPN module, and the decoupled detection head.
First, the Darknet backbone is employed to extract multi-resolution feature maps, incorporating DC to effectively adapt to defects of various shapes and sizes. Next, the DFPN module generates and concatenates multi-level feature maps, thereby enhancing the semantic richness of the extracted features. During forward propagation, network parameters are optimized through loss calculation and back-propagation. Finally, the decoupled detection head is responsible for target localization and classification, utilizing dense skip connections to fuse multi-scale features. This ensures the precise detection of surface defects in recycled and recirculated books.
This architecture enables DDNet to effectively address the challenges of detecting defects with varying shapes, sizes, and locations, delivering robust and accurate results.

\subsection{Feature Extraction via Deformable Convolution Operator }
In the detection of surface defects in recirculated books, it is a challenge to identify complex geometric deformations in images. Traditional methods such as SIFT, ORB, DPM and HOG are limited by artificially designed features and have difficulty in handling complex geometric deformations. Although convolutional neural networks (CNNs) are powerful, their fixed geometric structure limits the modeling of geometric deformations.

To solve this problem, DC is used to replace part of ordinary convolution. For tiny surface defects such as dirt, holes, wrinkles or cracks, traditional convolution kernels may lose key feature information. Deformable convolution learns the offset size by adding an offset field, allowing dynamic adjustment of the sampling position, thereby improving the flexibility and accuracy of feature extraction.

\begin{figure}[t]
 \vspace{-0.5cm} 
\centerline{\includegraphics[trim={0 10cm 0 0}, 
        clip,width=1\linewidth]{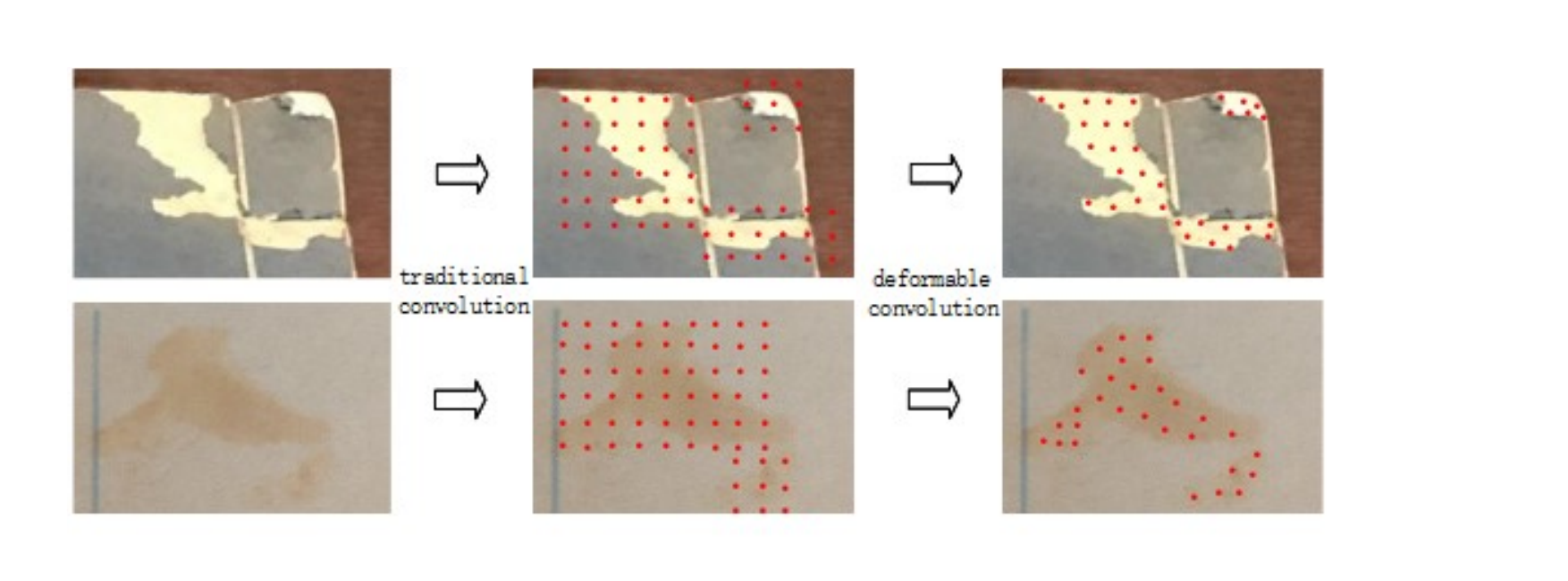}}
 \vspace{-0.5cm} 
\caption{Schematic diagram of feature calculation of ordinary convolution and deformable convolution.}
         \vspace{-0.6cm} 
\label{fig:F3}
\end{figure}

The calculation formula of deformable convolution is as follows:
\begin{equation}
    y(p_0) = \sum_{p_n \in R} \omega(p_n) \cdot x(p_0 + p_n + \Delta p_n)
    \label{eq:eq1}.
\end{equation}
Where, \(\Delta p_n\) is the offset, which enables the convolution operation to better adapt to changes in target shape and size.
 Fig.~\ref{fig:F3} shows the comparison between traditional convolution and deformable convolution with a 3×3 convolution kernel. The sampling position of traditional convolution is fixed, while deformable convolution can dynamically adjust its receptive field, improving robustness and adaptability, and is particularly suitable for processing complex and irregular target features. This technology plays a key role in the detection of surface defects in recirculated books, and can effectively identify and process defect features of different shapes and sizes.

\subsection{FPN with Dense Connections }
In convolutional neural networks, feature maps become increasingly abstract from the bottom to the top layers, with higher layers containing richer semantic information. Integrating high-level information back to lower layers can enhance multi-scale feature extraction and improve defect detection across various scales. Given the diverse scale and shape of defects, leveraging multi-scale features is crucial for robust and accurate detection. However, the Path Aggregation Feature Pyramid Network (PAFPN) used in YOLOv8 has limitations in feature reuse and multi-scale fusion.

To overcome these limitations, we propose a Dense Feature Pyramid Network (DFPN) with a dense connection strategy. DFPN connects feature maps at each scale with others from both higher and lower levels, enriching feature maps for subsequent detection stages.   Fig.~\ref{fig:F2} (b) illustrates the DFPN structure, which enhances feature transfer and improves defect identification in old book images.

In DFPN, feature maps at a specific level in the network are classified as feature layers of the same stage. We use C\textsubscript{*} (C\textsubscript{1}\textendash{}C\textsubscript{5}) to represent the feature maps of different sizes generated by the backbone network (CSPDarkNet-53); P\textsubscript{*} (P\textsubscript{1}\textendash{}P\textsubscript{5}) to represent the feature layers integrated through operations such as lateral connection and upsampling; and N\textsubscript{*} (N\textsubscript{1}\textendash{}N\textsubscript{5}) to represent the enhanced feature layers obtained by dense connection through the proposed DFPN.

Based on C\textsubscript{1} to C\textsubscript{5}, DFPN first performs a 1×1 convolution operation on C\textsubscript{i} to ensure the consistency of the number of channels in the fusion feature layer, and then fuses it with the upsampled feature map of C\textsubscript{i+1} by element-wise addition to generate a new feature map P\textsubscript{i}. Then, the P\textsubscript{i-1} layer feature is downsampled and connected to the P\textsubscript{i} layer feature map by tensor to obtain the preliminary fused feature map N\textsubscript{i}. Finally, in order to further improve the expressiveness of the features, DFPN fuses the generated N\textsubscript{i} feature map with all previous feature maps N\textsubscript{1}\textendash{}N\textsubscript{i-1} by dense connection. Specifically, for each feature map N\textsubscript{1}\textendash{}N\textsubscript{i-1}, use the scale of other feature maps downsampled to the same spatial size as N\textsubscript{i}, and then merge them by tensor concatenation to generate a richer feature map and send it to the detection head.

\section{EXPERIMENTS}
\subsection{Dataset DescriptionDataset}

The quality of images significantly influences the performance of surface defect detectors for recycled and recirculated books. To address the challenges posed by factors such as environment, lighting, noise, and camera variations, we developed the SHBD dataset, the first dataset focused on surface defect detection in these books. This dataset supports classification, object detection, and semantic segmentation tasks.

The SHBD dataset includes 6,366 images annotated for training, validation, and testing. It features six primary defect types: Vitium refers to missing book edges, corners or pages; Defaced may be water stains or oil stains; Crease indicates creases/cracks, a dog-ear or crack in a book page; Patch involves the use of adhesive or tape to repair a portion; Signature includes a collection, library, or sales seal; Inkiness includes a signature, note, or other trace of ink.

\begin{figure}[t]
 \vspace{-0.7cm} 
\centerline{\includegraphics[trim={2cm 5cm 0 0}, 
        clip,width=1\linewidth]{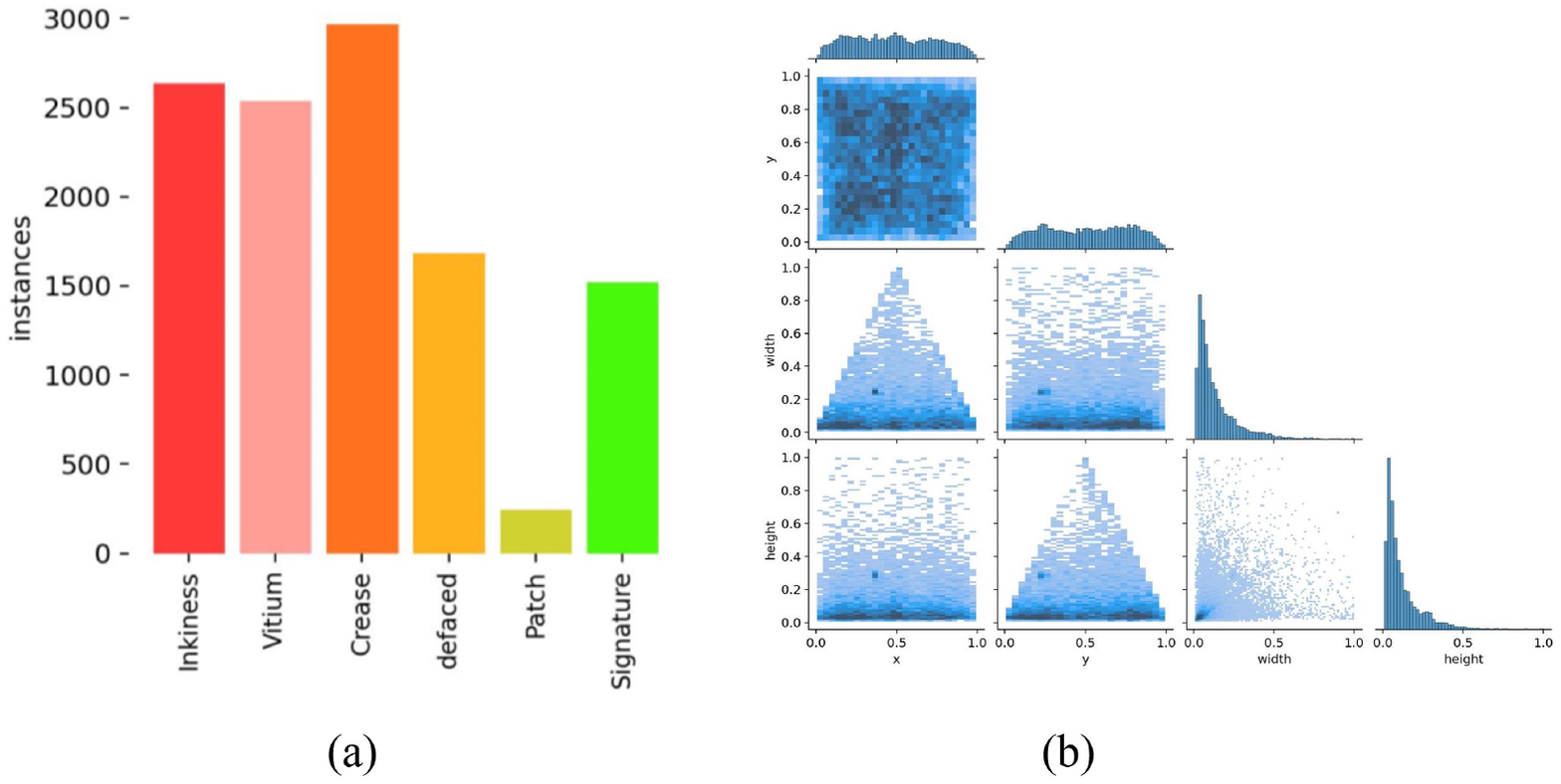}}
 \vspace{-0.6cm} 
\caption{Data feature analysis. (a)Histogram distribution of six categories of defects in SHBD. (b)Width and height distribution of all defects.}
 \vspace{-0.65cm} 
 \label{fig:F4}
\end{figure}

Defects may appear on various parts of the book, including covers and pages, and images can contain multiple defects. Analysis of this dataset provides insights into defect characteristics and advances inspection technology.   Fig.~\ref{fig:F1}  (d) shows typical defect images, and Fig.~\ref{fig:F4}  presents the defect proportion and size distribution, highlighting the challenges of defect diversity. The DC and DFPN in DDNet are specifically designed for these data features.

\subsection{Ablation experiment }  
Table ~\ref{tab:1}  summarizes the results of each phase of this study . Initially, we compared the performance of YOLOv5\cite{b12}, YOLOv6\cite{b13}, YOLOv8\cite{b14}, YOLOX\cite{b15} and rtm\cite{b16}  for surface defect detection in recycled books, selecting YOLOv8 as the baseline model with an mAP of 32.5\%.  In the second and third stages, two improved modules were introduced: DC and DFPN, respectively, to achieve different improvements.

In the second stage, introducing DC improved the mAP to 46.0\%, significantly enhancing detection accuracy for "Inkiness," "Crease," and "Defaced" categories by better capturing subtle shape variations. This shows that the added DC module is able to capture the subtle shape changes unique to these categories by dynamically adjusting the convolution grid to better match the object contour.

In the third stage, incorporating DFPN increased the mAP to 45.5\%, notably improving the "Patch" category from 19.4\% to 52.7\% and achieving a 94.0\% average accuracy for the "Signature" category.   This shows that DFPN enables the model to capture more details at different scales when dealing with defect classes with large size variations.

In the final stage, combining DC and DFPN to form the DDNet model improved the overall mAP by 14.2\% to 46.7\%, with a significant improvement in "patch" detection (mAP of 55.2\%). However, these improvements are not consistent across all defect classes, suggesting that optimizations for some defects may have a negative impact on others.

\begin{figure}[t]
    \vspace{-0.1cm} 
\centerline{\includegraphics[trim={0 6cm 0 0}, 
        clip,width=1\linewidth]{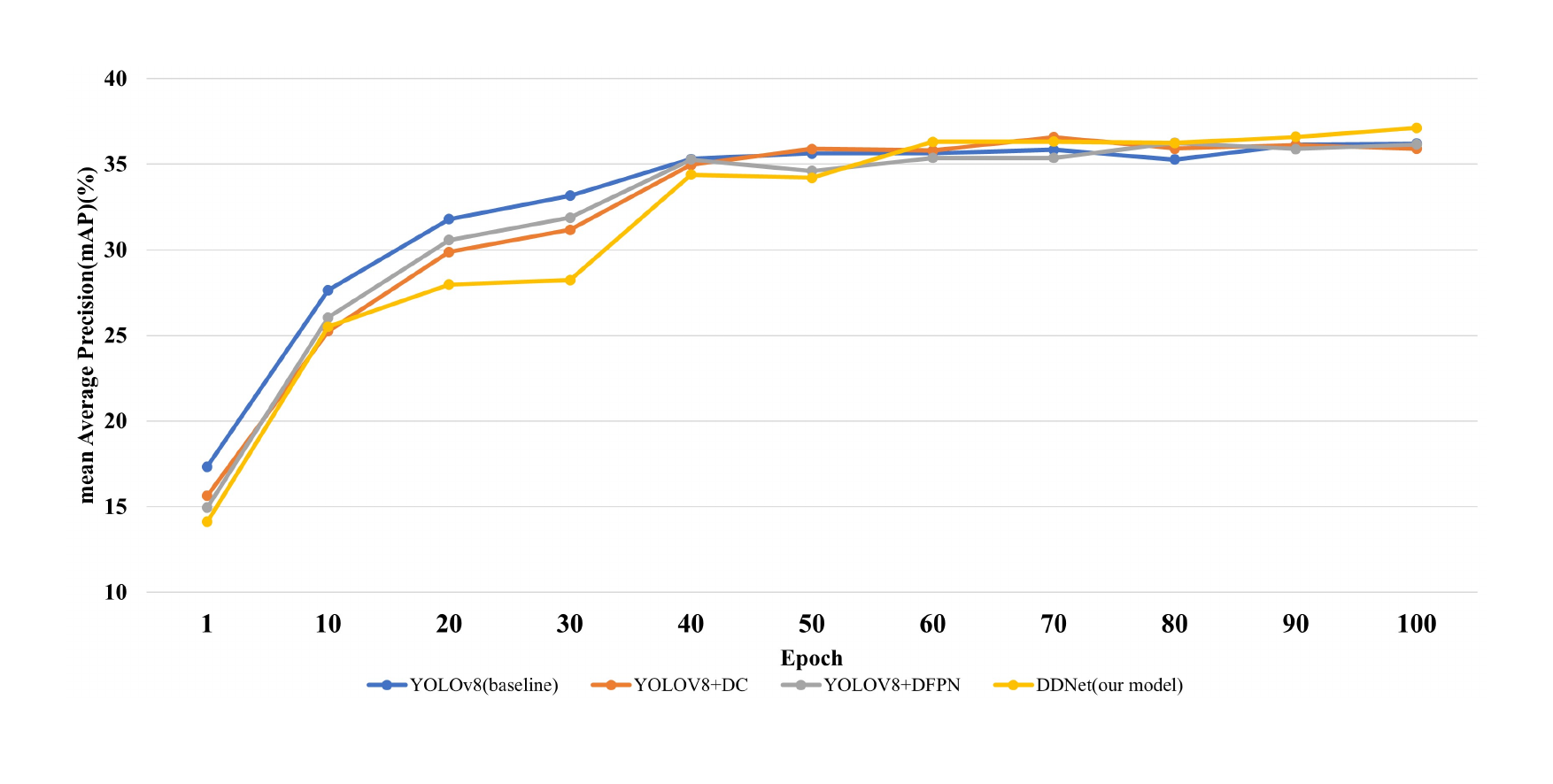}}
 \vspace{-0.6cm} 
\caption{Training accuracy over epochs of different improved methods .}
 \vspace{-0.7cm} 
 \label{fig:F5}
\end{figure}

Fig. \ref{fig:F5}  shows the performance of the models at different optimization stages. DDNet achieves the highest accuracy and the fastest convergence speed, outperforming other models in the later stages and continuing to improve. This confirms the effectiveness of the deep convolution (DC) and deep feature pyramid network (DFPN) methods used.

\begin{figure}[t]
    \vspace{-0.7cm} 
\centerline{\includegraphics[trim={0 9cm 0 0}, 
        clip,width=1\linewidth]{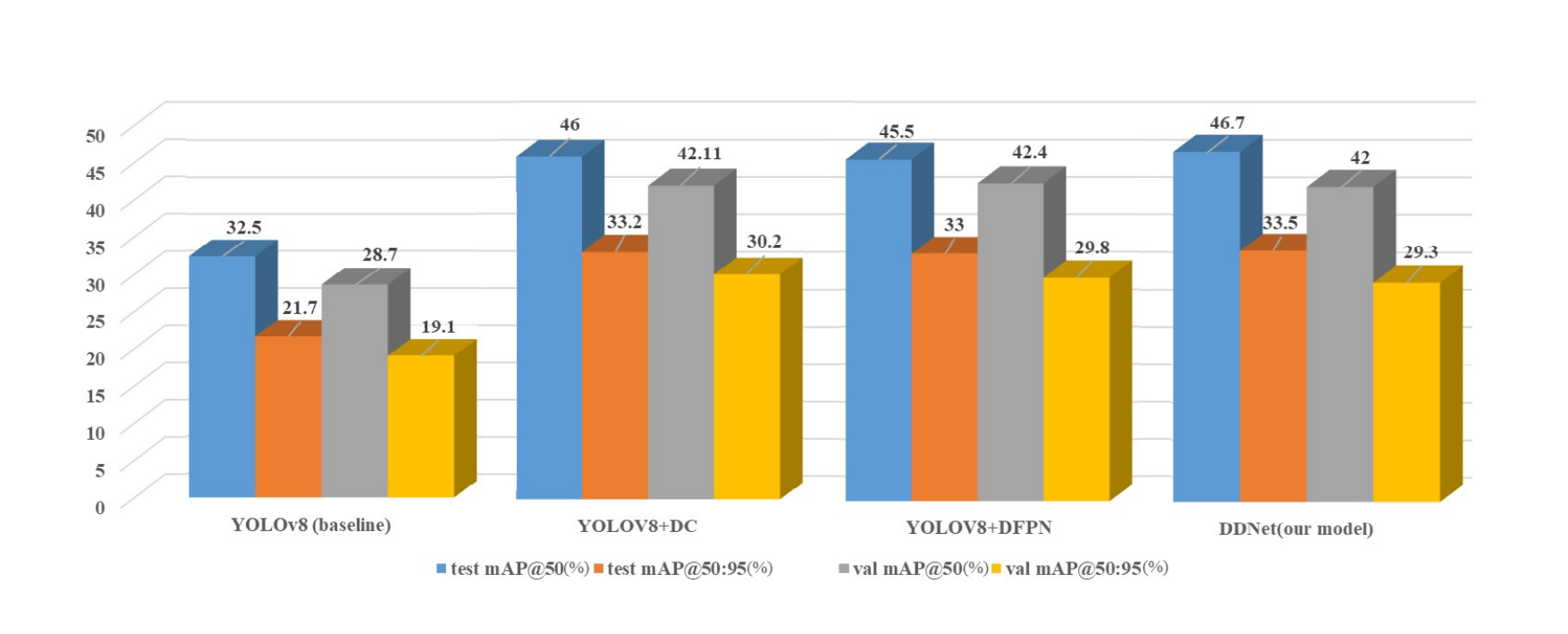}}
 \vspace{-0.6cm} 
\caption{Ablation study results.}
 \vspace{-0.3cm} 
 \label{fig:F6}
\end{figure}

Fig. \ref{fig:F6}   shows mAP results across four experimental phases. The DDNet model achieved mAP@50 and mAP@50:90 scores of 42.0\% and 29.3\% on the validation set, and 46.7\% and 33.5\% on the test set, the highest among the tested models. This highlights DDNet's superior overall performance, excellent localization capabilities, and robustness across different object scales and occlusion levels.

\subsection{Discussion }  

\paragraph{Optimization of adding DC layers to the backbone }

To enhance the network’s adaptability to various surface defects in reused books, we integrate the DC module into the CSPDarkNet backbone of YOLOv8. The results in Table ~\ref{tab:2}  show that the best detection performance is achieved when the DC module is added in the last two stages, achieving a balance between improving model adaptability and minimizing computational overhead.

\paragraph{The challenge of accurately detecting defects}

Although the model has achieved significant progress in overall performance, there are still issues with the detection accuracy of certain defects. 

\begin{figure}[t]
 \vspace{-0.45cm} 
\centerline{\includegraphics[trim={0 9cm 0 0}, 
        clip,width=1\linewidth]{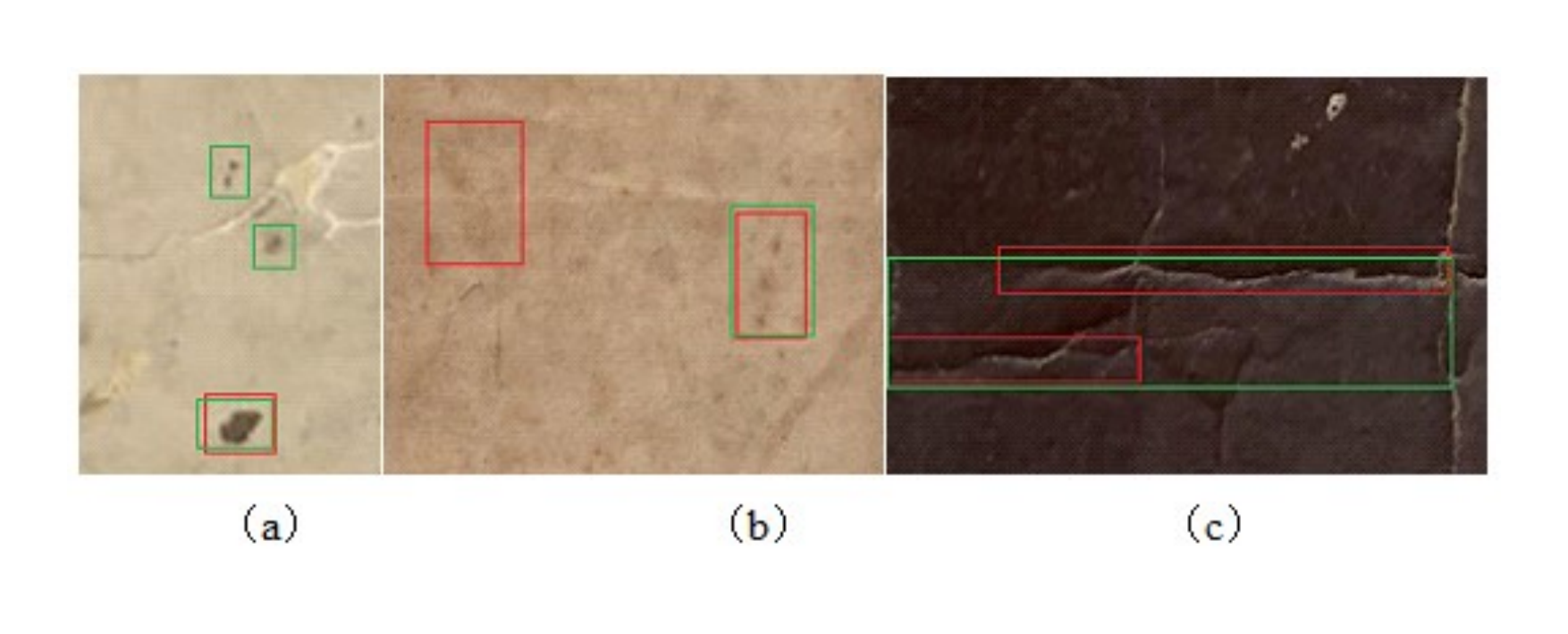}}
 \vspace{-0.8cm} 
\caption{Examples of inaccurate detection: Red boxes indicate ground truth, green boxes show detection results. (a) ‘Defaced’ defects blending with background. (b) ‘Crease’ defects with unclear borders. (c) Missing labels. }
         \vspace{-0.6cm} 
\label{fig:F7}
\end{figure}

\textbf{Low contrast between the defect and the background} (Fig. \ref{fig:F7} (a) ) complicates detection, indicating a need for improved handling of low-contrast features in the model.

\textbf{Blurred boundaries between adjacent defects  }(Fig. \ref{fig:F7} (b) ) often lead to misidentification of multiple defects as a single one. This suggests that the model may need further optimization when dealing with blurred boundaries or overlapping defects.

\textbf{Inconsistencies in dataset annotations } (Fig. \ref{fig:F7} (c) ) , such as missing labels for detected defects, can affect training effectiveness and overall model performance.

\section{Conclusion}
This study introduces DDNet, a novel model for detecting surface defects in recycled and recirculated books, utilizing deformable convolutions and dense skip connections to improve feature extraction and boundary accuracy. DDNet's precise localization and classification capabilities are demonstrated by a 14.2\% improvement in mAP over baseline models. Future work will aim to enhance model generalization for rare defects, investigate advanced data augmentation and semi-supervised learning techniques, and assess DDNet's effectiveness in real-world book recycling and sales scenarios .

\bibliographystyle{unsrt} 
\bibliography{references}

\end{document}